\documentclass{article}
\usepackage{courier}

\usepackage{ijcai24}

\usepackage{times}
\usepackage{soul}
\usepackage{url}
\usepackage[hidelinks]{hyperref}
\usepackage[utf8]{inputenc}
\usepackage[small]{caption}
\usepackage{graphicx}
\usepackage{amsmath}
\usepackage{amsthm}
\usepackage{booktabs}
\usepackage{algorithm}
\usepackage{algorithmic}
\usepackage[switch]{lineno}
\usepackage{color} 
\usepackage{amssymb}
\usepackage{amsmath}
\usepackage{listings}
\usepackage{tablefootnote}
\newcommand{\ttt}[1]{\texttt{\footnotesize #1}}

\usepackage{xspace}
\newcommand{\rb}{\textit{Combined Recurrency Baseline}\xspace}
\newcommand{\srb}{\textit{Strict Recurrency Baseline}\xspace} 
\newcommand{\rrb}{\textit{Relaxed Recurrency Baseline}\xspace}
\newcommand{\drec}{{direct recurrency degree}\xspace}
\newcommand{\rec}{{recurrency degree}\xspace}
\newcommand{\tp}{{t^+}\xspace}
\newcommand{\qt}{$(s, r, ?, \tp)$\xspace}
\newcommand{\qh}{$(?, r, o, \tp)$\xspace}
\newcommand{\G}{$G$\xspace}
\newtheorem{proposition}{Proposition}
\newtheorem{definition}{Definition}

\title{History Repeats Itself: A Baseline for Temporal Knowledge Graph Forecasting}
\author{
Julia Gastinger$^{1,2}$
\and
Christian Meilicke$^2$\and 
Federico Errica$^1$\and \\
Timo Sztyler$^1$ \and 
Anett Schuelke$^1$ \and
Heiner Stuckenschmidt$^2$
\affiliations
$^1$ NEC Laboratories Europe  \\
$^2$ University of Mannheim\\
\emails
\{julia.gastinger, federico.errica, timo.sztyler, anett.schuelke\}@neclab.eu\\
\{christian.meilicke, heiner.stuckenschmidt\}@uni-mannheim.de 
}

\begin{document}
\maketitle
\begin{abstract}
Temporal Knowledge Graph (TKG) Forecasting aims at predicting links in Knowledge Graphs for future timesteps based on a history of Knowledge Graphs. 
To this day, standardized evaluation protocols and rigorous comparison across TKG models are available, but the importance of simple baselines is often neglected in the evaluation, which prevents researchers from discerning actual and fictitious progress.
We propose to close this gap by designing an intuitive baseline for TKG Forecasting based on predicting recurring facts. Compared to most TKG models, it requires little hyperparameter tuning and no iterative training. Further, it can help to identify failure modes in existing approaches.
The empirical findings are quite unexpected: compared to 11 methods on five datasets, our baseline ranks first or third in three of them, painting a radically different picture of the predictive quality of the state of the art. 
\end{abstract}

\section{Introduction}
\label{sec:intro}
The lack of experimental rigor is one of the most problematic issues in fast-growing research communities, producing empirical results that are inconsistent or in disagreement with each other. Such ambiguities are often hard to resolve in a short time frame, and they eventually slow down scientific progress. 
This issue is especially evident in the machine learning field, where missing experimental details, the absence of standardized evaluation protocols, and unfair comparisons make it challenging to discern true advancements from fictitious ones \cite{lipton2019troubling}.

As a result, researchers have spent considerable effort in re-evaluating the performances of various models on different benchmarks, to establish proper comparisons and robustly gauge the benefit of an approach over others. In recent years, this was the case of node and graph classification benchmarks \cite{shchur_pitfalls_2018,errica_fair_2020}, link prediction on Knowledge Graphs \cite{Sun2020reevaluation,rossi2021knowledge}, neural recommender systems \cite{dacrema_are_2019}, and temporal graph learning \cite{huang2023temporal}.

Not only does such fast growing literature impact  reproducibility and replicability, but it is also characterized by a certain forgetfulness that simple baselines set a threshold above which approaches are actually useful. 
Oftentimes, these baselines are missing from the empirical evaluations, but when introduced they provide a completely new picture of the state of the art. Examples can be found in the field of Knowledge Graph completion, where simple rule-based systems can outperform embedding-based ones \cite{meilicke2018fine}, or in graph-related tasks where structure-agnostic baselines can compete with deep graph networks \cite{errica_fair_2020,poursafaei_towards_2022,errica2023class}.

In the last few years, the field of Temporal Knowledge Graph (TKG) Forecasting has also experienced a fast-paced research activity culminating in a large stream of works and a variety of empirical settings \cite{Liu2021tlogic,Sun21TimeTraveler,Zhang2023L2TKG}. Researchers have already provided a thorough re-assessment of some TKG Forecasting methods to address growing concerns about their reproducibility, laying down a solid foundation for future comparisons \cite{gastinger2023eval}. What is still missing, however, is a comparison with simple baselines to gauge if we are really making progress and to identify pain points of current representation learning approaches for TKGs.

Our contribution aims at filling this gap with a novel baseline, 
which places a strong inductive bias on the re-occurrence of facts over time. 
Not only does our baseline require tuning of just two hyperparameters, but also no training phase is needed since it is parameter-free. 
We introduce three variants of the baseline, divided into strict recurrency, relaxed recurrency, and a combination of both. 
Our empirical results convey an unexpected message: the baseline ranks first and third on three out of five datasets considered, compared to 11 TKG methods. It is a perhaps unsurprising result, given the long history of aforementioned works that propose strong baselines in different communities, but it further highlights the compelling need for considering simple heuristics in the TKG forecasting domain.  
Finally, by carefully comparing the performance of these baselines with other 
methods, we provide a failure analysis that highlights where it might be necessary to improve existing models.

\section{Related Work}
\label{sec:related}
In this section, we give a concise overview of the plethora of TKG forecasting methods that appeared in recent years. 

\paragraph{Deep Graph Networks (DGNs)} Several models in this category leverage message-passing architectures \cite{scarselli2009,micheli2009} along with sequential approaches to integrate structural and sequential information for TKG forecasting. RE-Net adopts an autoregressive architecture, learning temporal dependencies from a sequence of graphs \cite{Jin2020renet}. RE-GCN combines a convolutional DGN with a sequential neural network and introduces a static graph constraint to consider additional information like entity types \cite{Li2021regcn}.
xERTE employs temporal relational attention mechanisms to extract query-relevant subgraphs \cite{Han2021xerte}. TANGO utilizes neural ordinary differential equations and DGNs to model temporal sequences and capture structural information \cite{Han2021tango}.
CEN integrates a convolutional neural network capable of handling evolutional patterns in an online setting, adapting to changes over time \cite{Li2022cen}. 
At last, RETIA generates twin hyperrelation subgraphs and aggregates adjacent entities and relations using a graph convolutional network \cite{Liu2023retia}.

\paragraph{Reinforcement Learning (RL)} Methods in this category combine reinforcement learning with temporal reasoning for TKG forecasting. CluSTeR employs a two-step process, utilizing a RL agent to induce clue paths and a DGN for temporal reasoning \cite{Li2020cluster}. Also, TimeTraveler leverages RL based on temporal paths, using dynamic embeddings of the queries, the path history, and the candidate actions to sample actions, and a time-shaped reward \cite{Sun21TimeTraveler}.

\paragraph{Rule-based} Rule-based approaches focus on learning temporal logic rules. TLogic learns these rules via temporal random walks \cite{Liu2021tlogic}. TRKG extends TLogic by introducing new rule types, including acyclic rules and rules with relaxed time constraints \cite{Kiran2023TRKG}. ALRE-IR combines embedding-based and logical rule-based methods, capturing deep causal logic by learning rule embeddings \cite{Mei2022alreir}. LogE-Net combines logical rules with RE-GCN, using them in a preprocessing step for assisting reasoning \cite{Liu2023logenet}. At last, TECHS incorporates a temporal graph encoder 
and a logical decoder for differentiable rule learning and reasoning \cite{Lin2023TECHS}. 

\paragraph{Others} There are additional approaches with mixed contributions 
that cannot be immediately placed in the above categories. CyGNet predicts future facts based on historical appearances, employing a "copy" and "generation" mode~\cite{Zhu2021cygnet}. TiRGN employs a local encoder for evolutionary representations in adjacent timestamps 
and a global encoder to collect repeated facts \cite{Li2022TiRGN}. 
CENET distinguishes historical and non-historical dependencies through contrastive learning and a mask-based inference process \cite{Xu2023CENET}.
Finally, L2TKG utilizes a structural encoder and latent relation learning module to mine and exploit intra- and inter-time latent relations \cite{Zhang2023L2TKG}.

\section{Approach}
\label{sec:approach}
This section introduces several baselines: We start with the \srb, before moving to its ``relaxed'' version, the \rrb, and, ultimately, a combination of the two, the so-called \rb. Before we introduce these baselines, we give a formal definition of the notion of a Temporal Knowledge Graph and and provide a running example to illustrate our approach.

\subsection{Preliminaries}
\label{sub:approach-prelim}
 A \textit{Temporal Knowledge Graph} \G  is a set of quadruples $(s,r,o,t)$ with $s, o \in E$, relation $r \in R$, and time stamp $t \in T$ with $T = \{ 1 \ldots n \}, n \in \mathbb{N}^+$. More precisely, $E$ is the set of entities, $R$ is the set of possible relations, and $T$ is the set of timesteps. 
 A quadruple's $(s,r,o,t)$ semantic meaning is that $s$ is in relation $r$ to $o$ at $t$. 
 Alternatively, we may refer to this quadruple as a temporal triple that holds during the timestep $t$.
 This allows us to talk about the triple $(s,r,o)$ and its occurrence and recurrence at certain timesteps. 
In the following, we use a running example \G, where \G is a TKG in the soccer domain shown in Figure~\ref{listing:roque}. \G contains triples from the years 2001 to 2009, which we map to indices 1 to 9.
\begin{figure}
\begin{lstlisting}
(marta, playsFor, vasco-da-gamah, 1)
(marta, playsFor, vasco-da-gamah , 2)
(marta, playsFor, santa-cruz, 3)
(marta, playsFor, santa-cruz, 4)
(marta, playsFor, umea-ik, 5)
(marta, playsFor, umea-ik, 6)
(marta, playsFor, umea-ik, 7)
(marta, playsFor, umea-ik, 8)
(marta, playsFor, los-angeles-sol, 9)
\end{lstlisting}
\caption{A (slightly simplified) listing of the clubs that Marta Vieira da Silva, known as Marta, played for from 2001 to 2009.}
\label{listing:roque}
\end{figure}

\textit{Temporal Knowledge Graph Forecasting} is the task of predicting quadruples for future timesteps $\tp$ given a history of quadruples \G, with $\tp > n$ and $\tp \in \mathbb{N}^+$. 
In this work we focus on entity forecasting, that is, predicting object or subject entities for queries $(s,r,?,\tp)$ or $(?,r,o,\tp)$. Akin to KG completion, TKG forecasting is approached as a ranking task \cite{han2022TKG}. For a given query, e.g. \qt, methods rank all entities in $E$ using a scoring function, assigning plausibility scores to each quadruple. 

In the following, we design several variants of a simple scoring function $f$ that assigns a score in $\mathbb{R^+}$ to a quadruple at a future timestep $\tp$ given a Temporal Knowledge Graph \G, i.e., $f((s,r,o,\tp), G) \mapsto \mathbb{R^+}$. 
All variants of our scoring function are simple heuristics to solve the TKG forecasting task, based on the principle that \emph{something that happened in the past will happen again in the future}.

\subsection{Strict Recurrency Baseline}
\label{sub:approach-recurrency}
The first family of recurrency baselines checks if the triple that we want to predict at timestep $\tp$ has already been observed before. 
The simplest baseline of this family is the following scoring function $\phi_1$:
\begin{equation}
\phi_1((s,r,o,\tp), G) =
\begin{cases}
  1, & \text{if}\ \exists k \ \text{with} \ (s,r,o,k) \in G \\
  0, & \text{otherwise.}
\end{cases}
\end{equation}

If we apply $\phi_1$ to the set of triples in Figure~\ref{listing:roque} to compute the scores for 2010, we get the following outcome (using \ttt{pf} to abbreviate \ttt{playsFor}). 
\begin{align*}
& \phi_1((\ttt{marta}, \ttt{pf}, \ttt{vasco-da-gamah}, \ttt{10}), G) = 1\\
& \phi_1((\ttt{marta}, \ttt{pf}, \ttt{santa-cruz}, \ttt{10}), G) = 1 \\
& \phi_1((\ttt{marta}, \ttt{pf}, \ttt{umea-ik}, \ttt{10}), G) = 1 \\
& \phi_1((\ttt{marta}, \ttt{pf}, \ttt{los-angeles-sol}, \ttt{10}), G) = 1
\end{align*}
This scoring function suffers from the problem that it does not take the temporal distance into account, which is highly relevant for the relation of playing for a club. It is far more likely that Marta will continue to play for Los Angeles Sol rather than sign a contract with a previous club. 

To address this problem, we introduce a time weighting mechanism to assign higher scores to more recent triples. Defining a generic function $\Delta: \mathbb{N}^+ \times \mathbb{N}^+ \rightarrow \mathbb{R}$ that takes the query timestep $\tp$, a previous timestep $k$ in $G$, and returns the weight of the triple,
we can define strict recurrency scoring functions as follows:
\begin{equation} \label{eq:phidelta}
\thickmuskip=0\thickmuskip
\thinmuskip=0\thinmuskip
\phi_\Delta((s,r,o,\tp), G) =
\begin{cases}
  \Delta(\tp, \max \{ k | (s,r,o,k) \in G \})  \\
  0, \ \ \ \text{if}\ \nexists k \ \text{with} \ (s,r,o,k) \in G.
\end{cases}
\end{equation}
For instance, using $\Delta^0(\tp,k) = k / \tp, k < \tp$ produces:
\small
\begin{align*}
& \phi_{\Delta^0}((\ttt{marta}, \ttt{pf}, \ttt{vasco-da-gamah}, \ttt{10}), G) = 0.2\\
& \phi_{\Delta^0}((\ttt{marta}, \ttt{pf}, \ttt{santa-cruz}, \ttt{10}), G) = 0.4 \\
& \phi_{\Delta^0}((\ttt{marta}, \ttt{pf}, \ttt{umea-ik}, \ttt{10}), G) = 0.8 \\
& \phi_{\Delta^0}((\ttt{marta}, \ttt{pf}, \ttt{los-angeles-sol}, \ttt{10}), G) = 0.9,
\end{align*}
\normalsize
which already makes more sense: the latest club that a person played for will always receive the highest score.

Interestingly, we can establish an equivalence class among a subset of the functions $\phi_\Delta$, and we will use this fact in our experiments. As long as we solely focus on ranking results, two scoring functions are equivalent if they define the same partial order over all possible temporal predictions.
\begin{definition}
\label{def:rank}
Two scoring functions $\phi$ and $\phi'$ are ranking-equivalent if for any pair of predictions $p = (s,r,o,\tp)$ and $p' = (s',r',o',\tp)$ we have that $\phi(p, G) > \phi(p', G) \iff \phi'(p, G) > \phi'(p', G)$.
\end{definition}
 
The next result states that we do not need to search for an optimal time weighting function $\Delta(\tp,k)$ if we choose it to be strictly monotonically increasing with respect to $k$, as these functions belong to the same equivalence class.
\begin{proposition}
\label{pro:phi-requiv}
Scoring functions $\phi_\Delta$ and $\phi_{\Delta'}$ are ranking equivalent iff, $\forall \ k_1,k_2,\tp$ such that $k_1 < k_2 < \tp$ it holds $\Delta(\tp,k_1) < \Delta(\tp,k_2)$ and $\Delta'(\tp,k_1) < \Delta'(\tp,k_2)$.
\end{proposition}
Proposition~\ref{pro:phi-requiv} follows from the application of Definition~\ref{def:rank}.
Therefore, the set of functions $\phi_{\Delta}$, characterized by a $\Delta$ that is strictly monotonically increasing in $k$, are ranking equivalent.

While $\phi_\Delta$ works well to predict the club that a person will play for, there are relations with different temporal characteristics. 
An example might be a relation that expresses that a soccer club wins a certain competition. In Figure~\ref{listing:bundesliga}, we extend our TKG with temporal triples using the relation wins.

\begin{figure}[t]
\begin{lstlisting}
(fc-bayern-munich, wins, bundesliga, 1)
(borussia-dortmund, wins, bundesliga, 2)
(fc-bayern-munich, wins, bundesliga, 3)
(werder-bremen, wins, bundesliga, 4)
(fc-bayern-munich, wins, bundesliga, 5)
(fc-bayern-munich, wins, bundesliga, 6)
(vfb-stuttgart, wins, bundesliga, 7)
(fc-bayern-munich, wins, bundesliga, 8)
(vfl-wolfsburg, wins, bundesliga, 9)
\end{lstlisting}
\caption{Clubs winning the Bundesliga 
from 2001 to 2009.}
\label{listing:bundesliga}
\end{figure}

The relation {wins} seems to follow a different pattern compared to the previous example. Indeed, applying $\phi_{\Delta^0}$ to predict the 2010 winner of the Bundesliga would not reflect the fact that FC Bayern Munich is the club with the highest ratio of won championships, and year 9 might just have been a lucky one for VFL Wolfsburg. The frequency of wins could be considered a better indicator for a scoring function: 
\begin{equation}
\label{eq:psi1}
\psi_1((s,r,o,\tp), G) = |\{ k \ | \ (s,r,o,k) \in G \}| / \tp
\end{equation}
Based on this scoring function, the club that has won the most titles, Bayern Munich, receives the highest score of $0.6$, while all other clubs receive a score of $0.1$.

As done earlier, we now generalize the formulation of $\psi_1$ to $\psi_\Delta$ using a weighting function $\Delta(\tp,k)$ where triples that occurred more recently are weighted higher:
\begin{equation}
\psi_\Delta((s,r,o,\tp), G) = \frac{\sum_{i \in \{ k \mid (s,r,o,k) \in G \}} \Delta(\tp,i)}{\sum_{i=1}^n \Delta(\tp,i)}.
\end{equation}
Again, we apply the new scoring functions to our example.
We shortened the names of the clubs and abbreviated \ttt{bundesliga} as \ttt{bl}:
\begin{align*}
& \psi_{\Delta^0}((\ttt{dortmund}, \ttt{wins}, \ttt{bl}, \ttt{10}), G) = 0.2/ 4.5 \approx 0.04 \\
& \psi_{\Delta^0}((\ttt{bremen}, \ttt{wins}, \ttt{bl}, \ttt{10}), G) =  0.4/ 4.5 \approx 0.09 \\
& \psi_{\Delta^0}((\ttt{stuttgart}, \ttt{wins}, \ttt{bl}, \ttt{10}), G) = 0.7 / 4.5 \approx 0.15 \\
& \psi_{\Delta^0}((\ttt{munich}, \ttt{wins}, \ttt{bl}, \ttt{10}), G) = 2.3 / 4.5 \approx 0.51 \\
& \psi_{\Delta^0}((\ttt{wolfsburg}, \ttt{wins}, \ttt{bl}, \ttt{10}), G) = 0.9 / 4.5 \approx 0.2 
\end{align*}
It is worth noting that, for a restricted family of distributions $\Delta'(t,k)$, we can achieve ranking equivalence between scoring functions $\psi_{\Delta'}$ and $\phi_\Delta$ with a strictly increasing $\Delta(t,k)$. More specifically, if we make $\Delta'(t,k)$ parametric, then $\psi_{\Delta'}$ can generalize the family of scoring functions $\phi_\Delta$.
Consider the parameterized function $\Delta_{\lambda}(\tp,k) = 2^{\lambda (k - \tp)}$ with $\lambda \in \mathbb{R}_0^+$, where $\lambda$ acts as a decay factor. 
The higher $\lambda$, the stronger the decay effect we achieve.
In particular, if we set $\lambda = 1$, we can enforce that a time point $k$ always receives a higher weight than the sum of all previous time points $1, \ldots, k-1$. This means $\psi_{\Delta_{1}}$ and $\phi_\Delta$ are ranking equivalent.

\begin{proposition}
\label{pro:lambda1}
For $\lambda\geq1$, $\Delta_\lambda = 2^{\lambda(k-\tp)}$, and any strictly increasing time weighting function $\Delta$, the scoring functions $\phi_\Delta$ and $\psi_{\Delta_{\lambda}}$ are ranking equivalent.
\end{proposition}
Proposition~\ref{pro:lambda1} follows directly from the fact that $\sum_{i=k+1}^n \frac{1}{2^i} < \frac{1}{2^{k}}$ for any $n > k \in \mathbb{N}^+$.

On the contrary, we get ranking equivalence between $\psi_1$ and $\psi_{\Delta_{\lambda}}$ if we set $\lambda=0$.
\begin{proposition}
\label{pro:lambda0}
The scoring functions $\psi_1$ and $\psi_{\Delta_{\lambda}}$ are ranking equivalent if we set $\lambda = 0$.
\end{proposition}
Proposition~\ref{pro:lambda0} follows directly from $2^0 = 1$ and the definition of $\psi_1$ in Equation~\ref{eq:psi1}.
Propositions~\ref{pro:lambda1} and~\ref{pro:lambda0} help us to interpret our experimental results, as it indicates that different settings of $\lambda$ result in a scoring function that is situated between $\psi_1$ and $\phi_{\Delta_{\lambda}}$. %
We treat $\lambda$ as a \textit{relation-specific} hyperparameter in our experiments, meaning we will select a different $\lambda_r$ for each relation $r$. Since relations are independent of each other, each $\lambda_r$ can be optimized independently.

\subsection{Relaxed Recurrency Baseline}
\label{sub:approach-frequency}
So far, our scoring functions were based on a strict application of the principle of recurrency. However, this approach fails to score a triple that has never been seen before, and we need to account for queries of this nature: 
imagine a young player appearing for the first time in a professional club.

Thus, we introduce a relaxed variant of the baseline. 
Instead of looking for exact matching of triples in previous timesteps, which would not work for unseen triples, we are interested in how often \textit{parts of the triple} have been observed in the data. When asked to score the query $(s,r,?,\tp)$, we compute the normalized frequency that the object $o$ has been in relationship $r$ with \textit{any} subject $s'$:
\begin{align}\label{eq:xi-obj}
\overrightarrow{\xi}((s,r,o,\tp), G) = \frac{|\{ (s',k) \ | \ (s',r,o,k) \in G \}|}{|\{ (s',o',k) \ | \ (s',r,o',k) \in G \}|}
\end{align}
Analogously, we denote with $\overleftarrow{\xi}((s,r,o,\tp), G)$ the relaxed baseline used to score queries of the form $(?,r,o,\tp)$. In the following, we omit the arrow above $\xi$ and use the directed version depending on the type of query without explicit reference to the direction. 

Let us revisit the example of Figure \ref{listing:roque} and apply $\xi$ to score a triple never seen before. 
We can now assign non-zero scores to the clubs that Aitana Bonmati, who never appeared in $G$, will likely play for in 2010:
\begin{align*}
& \xi((\ttt{bonmati}, \ttt{pf}, \ttt{vasco-da-gamah}, \ttt{10}), G) = 0.22 \\
& \xi((\ttt{bonmati}, \ttt{pf}, \ttt{santa-cruz}, \ttt{10}), G) = 0.22 \\
& \xi((\ttt{bonmati}, \ttt{pf}, \ttt{umea-ik}, \ttt{10}), G) = 0.44 \\
& \xi((\ttt{bonmati}, \ttt{pf}, \ttt{los-angeles-sol}, \ttt{10}), G) = 0.11
\end{align*}

While we also report results for $\xi$ on its own, we are mainly interested in its combination with the the \srb, where we expect it to fill up gaps and resolve ties. For simplicity, we do not introduce a weighted version of this baseline to avoid the extra hyperparameter.

\subsection{Combined Recurrency Baseline}
We conclude the section with  
a linear combination of the \srb $\psi_{\Delta_\lambda}$ and the \rrb $\xi$. In particular (omitting $\lambda$ to keep the notation uncluttered):
\begin{equation}
\begin{split}
    \psi_\Delta\xi((s,r,o,\tp), G)  = \alpha * \psi_{\Delta} (s,r,o,\tp), G) +\\ (1-\alpha) * \xi(s,r,o,\tp), G),
\end{split}
\end{equation}
where $\alpha \in [0,1]$ is another hyperparameter. Similar to $\lambda$, we select a different $\alpha_r$ for each relation $r$. In the following, we refer to this baseline as the \rb.



\section{Experimental Setup}
\label{sec:exp}

This section describes our experimental setup and provides information on how to reproduce our experiments\footnote{\url{https://github.com/nec-research/recurrency_baseline_tkg}.}. We rely on the unified evaluation protocol of \cite{gastinger2023eval}, 
reporting results about single-step predictions. We report results for the multi-step setting in the supplementary material\footnote{Supplementary Material: \url{https://github.com/nec-research/recurrency_baseline_tkg/blob/master/supplementary_material.pdf}}. 

\subsection{Hyperparameters}
We select the best hyperparameters 
by evaluating the performances on the validation set as follows:
First, we select $\lambda_r\forall r \in R$ from in total 14 values, $\lambda_r \in L_r=\{0,...,1.0001\}$ for $\psi_\lambda$. Then, after fixing the best $\lambda_r \forall r \in R$, we select $\alpha_r \forall r \in R$ from 13 values, $\alpha_r \in A_r=\{0,...,1\}$, leading to a total of 27 combinations per relation.


\subsection{Methods for Comparison}
We compare our baselines to 11 among the 17 methods described in Section~\ref{sec:related}. 
Two of these 17 methods run only in multi-step setting, see comparisons to these in the supplementary material.
Further, for four methods we find discrepancies in the evaluation protocol and thus exclude them from our comparisons\footnote{CENET, RETIA, and CluSTER do not report results in time-aware filter setting. ALRE-IR does not report results on WIKI, YAGO, and GDELT, and uses different dataset versions for ICEWS14 and ICEWS18.}.
Unless otherwise stated, we report the results for these 11 methods based on the evaluation protocol by \cite{gastinger2023eval}. For TiRGN, we report the results of the original paper and do a sanity check of the released code. We do the same for L2TKG, LogE-Net, and TECHS, but we cannot do a sanity check as their code has not been released. 

\subsection{Dataset Information}
We assess the performance of the recurrency baselines on five datasets \cite{gastinger2023eval,Li2021regcn}, namely WIKI, YAGO, ICEWS14, ICEWS18, and GDELT\footnote{See Supplementary Material for additional dataset information.}. 
Table~\ref{table:datasets} shows characteristics such as the number of entities and quadruples, and it reports the timestep-based data splitting (short: \#Tr/Val/Te TS) all methods are evaluated against.
In addition, we compute the fraction of test temporal triples $(s,r,o,\tp)$ for which there exists a $k<\tp$ such that $(s,r,o,k) \in G$, and we refer to this measure as the \textit{recurrency degree (Rec)}. Similarly, we also compute the fraction of temporal triples $(s,r,o,\tp)$ for which it holds that $(s,r,o,\tp-1) \in G$, which we call \textit{direct} recurrency degree \textit{(DRec)}. Note that \textit{Rec} defines an upper bound of \srb's performance; instead, \textit{DRec} informs about the test triples that have, from our baselines' perspective, a trivial solution. On YAGO and WIKI, both measures are higher than 85\%, meaning that the application of the recurrency principle would likely work very well. 
\begin{table*}[t]

\small
\centering
\begin{tabular}{lrrrrrrrrr}
\toprule
Dataset &  $\#$Nodes &  $\#$Rels& $\#$Train & $\#$Valid & $\#$Test & Time Int. & $\#$Tr/Val/Te TS   & DRec [\%] & Rec [\%]\\
\midrule
ICEWS14 &  $7128$ 	&  $230$ 	& $74845$	& $8514$	& $7371$	& 24 hours & 304/30/31          & 10.5 & 52.4
 \\
ICEWS18 &  $23033$ 	&  $256$ 	& $373018$ 	& $45995$ 	& $49545$ 	& 24 hours &  239/30/34         & 10.8 & 50.4
\\
GDELT &  	$7691$ 	&  $240$ 	& $1734399$& $238765$	& $305241$	& 15 min. & 2303/288/384        & 2.2 & 64.9
\\
YAGO &  	$10623$ 	&  $10$ 	& $161540$	& $19523$	& $20026$	& 1 year & 177/5/6          & 92.7 & 92.7
\\
WIKI &  	$12554$ 	&  $24$ 	& $539286$	& $67538$	& $63110$	& 1 year & 210/11/10        & 85.6 & 87.0
\\

\bottomrule
\end{tabular}
\small
\caption{We report some statistics of the datasets, the timestep interval, and the specifics of the data splitting. We also include the {recurrency degree (Rec)} and the {direct recurrency degree (DRec)}. Please refer to the text for a more detailed description.}
\label{table:datasets}
\end{table*}

\subsection{Evaluation Metrics}
As is common in link prediction evaluations, we focus on two metrics: the Mean Reciprocal Rank (MRR), computing the average of the reciprocals of the ranks of the first relevant item in a list of results, as well as the Hits at 10 (H@10), the proportion of queries for which at least one relevant item is among the top 10 ranked results. Following \cite{gastinger2023eval}, we report the time-aware filtered MRR and H@10.

%

\section{Experimental Results}
\label{sec:results}
This section reports our quantitative and qualitative results, illustrating our baselines help to gain a deeper understanding of the field. We list runtimes in the Supplementary Material.

\subsection{Global Results}\label{sec:sota}
\setlength{\tabcolsep}{5pt}
\begin{table*} 
\small
\centering
\begin{tabular}{l|rr|rr|rr|rr|rr}
\toprule
& \multicolumn{2}{c|}{GDELT} & \multicolumn{2}{c|}{YAGO} & \multicolumn{2}{c|}{WIKI} & \multicolumn{2}{c|}{ICEWS14} & \multicolumn{2}{c}{ICEWS18} \\
\midrule
		                  & MRR  & H@10 & MRR  & H@10 & MRR  & H@10 & MRR & H@10 & MRR & H@10  \\
\midrule
L2TKG$^{\dagger}$     & 20.5          & 35.8          & -             & -             & -             & -             & \textbf{47.4} & \textbf{71.1} & \ul{{33.4}} & \textbf{55.0} \\
LogE-Net$^{\dagger}$  & -             & -             & -             & -             & -             & -             & 43.7          & 63.7          & 32.7                & 53.0          \\
TECHS$^{\dagger}$     & -             & -             & 89.2          & 92.4          & 76.0          & 82.4          & d.d.v         & d.d.v.        & 30.9                & 49.8          \\
TiRGN                  & 21.7          & 37.6          & 88.0          & \ul{92.9}     & \ul{81.7}     & \textbf{87.1} & \ul{44.0}     & \ul{63.8}     & \textbf{33.7}       & \ul{54.2}    \\
TRKG                   & 21.5          & 37.3          & 71.5          & 79.2          & 73.4          & 76.2          & 27.3          & 50.8          & 16.7                & 35.4          \\
RE-GCN                 & 19.8          & 33.9          & 82.2          & 88.5          & 78.7          & 84.7          & 42.1          & 62.7          & 32.6                & 52.6          \\
xERTE                  & 18.9          & 32.0          & 87.3          & {91.2}        & 74.5          & {80.1}        & 40.9          & 57.1          & 29.2                & 46.3          \\
TLogic                 & 19.8          & 35.6          & 76.5          & 79.2          & \textbf{82.3} & 87.0          & 42.5          & 60.3          & 29.6                & 48.1          \\
TANGO                  & 19.2          & 32.8          & 62.4          & 67.8          & 50.1          & 52.8          & 36.8          & 55.1          & 28.4                & 46.3          \\
Timetraveler & 20.2          & 31.2          & 87.7          & 91.2          & 78.7          & 83.1          & 40.8          & 57.6          & 29.1                & 43.9          \\
CEN                    & 20.4          & 35.0          & 82.7          & 89.4          & 79.3          & 84.9          & 41.8          & 60.9          & 31.5                & 50.7          \\
\midrule
\textit{Relaxed} ($\xi$)         & 14.2          & 23.6          & 5.2           & 10.7          & 14.3          & 25.4          & 14.4          & 28.6          & 11.6                & 22.0          \\
\textit{Strict} ($\psi_\Delta$)  & \ul{23.7}    & \ul{38.3}    & \ul{90.7}    & 92.8          & 81.6          & 87.0          & 36.3          & 48.4          & 27.8                & 41.4          \\
\textit{Combined} ($\psi_\Delta\xi$)      & \textbf{24.5} & \textbf{39.8} & \textbf{90.9} & \textbf{93.0} & 81.5          & \textbf{87.1} & 37.2          & 51.8          & 28.7                & 43.7          \\ 
\bottomrule
  \end{tabular}
  \caption{Experimental results. An entry ${\dagger}$ means authors have not released their code, and thus we could not reproduce their results, an entry~- that the related work does not report results on this dataset, and an entry "d.d.v", that the it reports results on a different dataset version.}
\label{table:results}
\end{table*}
Table \ref{table:results} (lower area) shows the MRR and H@10 results for the \textit{Strict} ($\xi$), the \textit{Relaxed} ($\psi_\Delta$), and the \rb~($\psi_\Delta\xi$). 
For all datasets, with one minor discrepancy, the \rb performs better than the strict and the relaxed variants. However, the \srb is not much worse: The difference to the \rb is for both metrics never more than one percentage point. We observe that, while $\xi$ scores a MRR between $5\%$ and $15\%$ on its own, when combined with $\psi_\Delta$ (thus obtaining $\psi_\Delta\xi$) it can grant up to $0.9\%$ of absolute improvement. 
As described in Section \ref{sec:approach}, its main role is to fill gaps and resolve ties. The results confirm our intuition.
Interestingly, results for $\psi_\Delta\xi$ on all datasets reflect the reported values of the \rec and \drec (see Table~\ref{table:results}): For both YAGO and WIKI (Rec and DRec~$>85\%$), 
our baseline yields high MRRs~($>80\%$), while in other cases the values are below~$40\%$. 

When compared to results from related work (upper area of Table~\ref{table:results}), 
the \rb as well as the \srb yield the highest test scores for two out of five datasets (GDELT and YAGO) and the third-highest test scores for the WIKI dataset.
This is an indication that most related work models seem unable to learn and consistently apply a simple forecasting strategy that yields high gains. 
In particular, we highlight the significant difference between the \rb and the runner-up methods for GDELT (with a relative change of $+12.9\%$). 

Results for ICEWS14 and ICEWS18, instead, suggest that more complex dependencies need to be captured on these datasets. While two methods (TRKG and TANGO) perform worse than our baseline, the majority achieves better results. 

In summary, none of the methods proposed so far can accomplish the results achieved by a combination of two very naïve baselines for two out of five datasets.
This result is rather surprising, and it raises doubts about the predictive quality of current methods.

\subsection{Per-Relation Analysis}\label{sec:per-rel}
We conduct a detailed per-relation analysis and focus on two datasets:
ICEWS14, since our baseline performed worse there, and YAGO, for the opposite reason. We compare the \rb to the four methods that performed best on the respective dataset, considering the seven methods evaluated under the evaluation protocol of \cite{gastinger2023eval}\footnote{Since we could compute prediction scores for every query.}. 
For clarity, we adopt the following notation to denote a relation and its prediction direction: {[relation] (head)} signifies predictions in head direction, corresponding to queries of the form \qh; {[relation] (tail)} denotes predictions in tail direction, i.e., \qt. 

\begin{figure}
\begin{minipage}{.48\textwidth}
\centering
\small{(a) ICEWS14}
\vspace{0.1cm}
\includegraphics[width=\columnwidth]{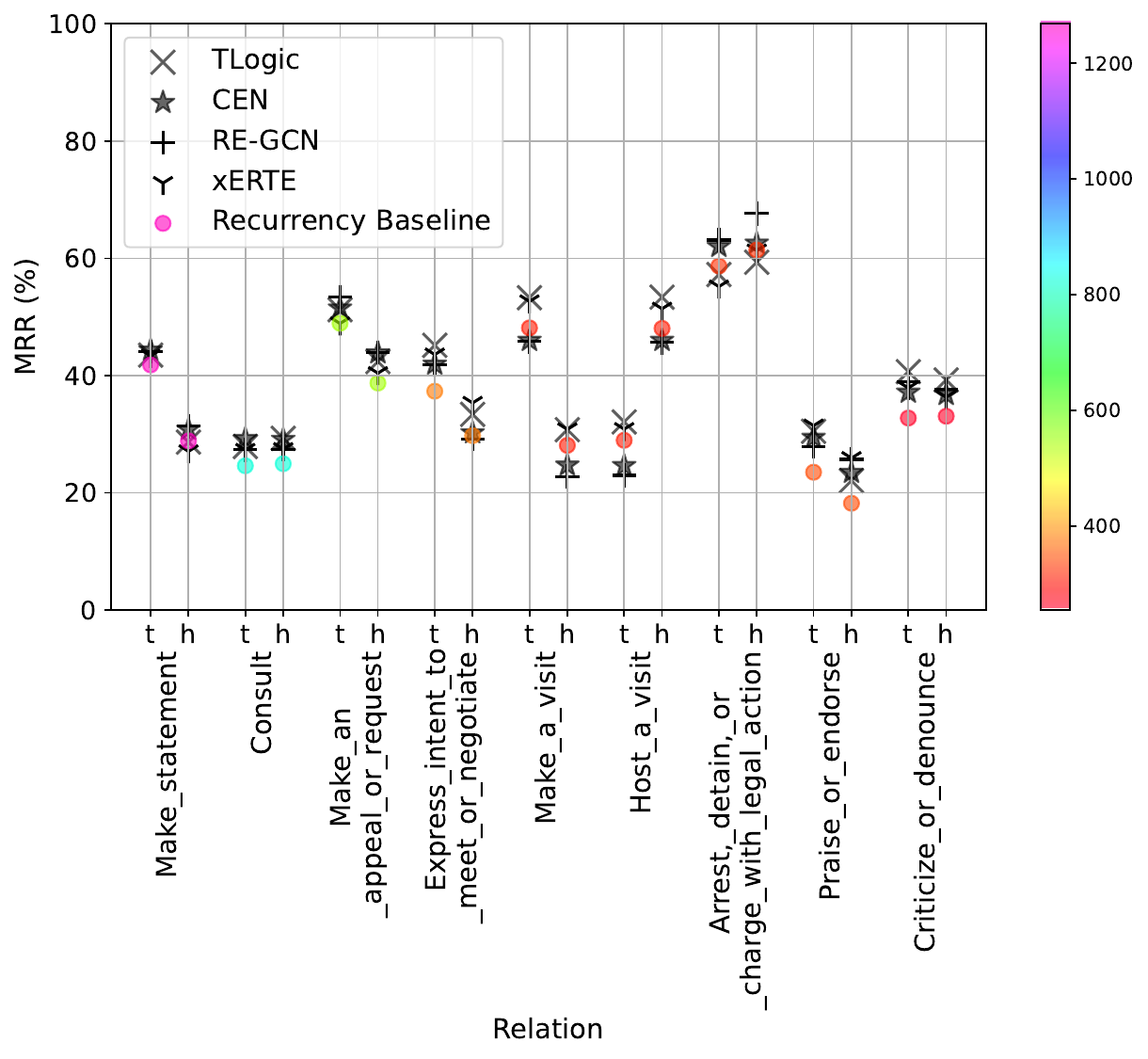}
\end{minipage}
\begin{minipage}{.48\textwidth}
\centering
\vspace{0.5cm}
\small{(b) YAGO}
\vspace{0.1cm}
 \centering
\includegraphics[width=\columnwidth]{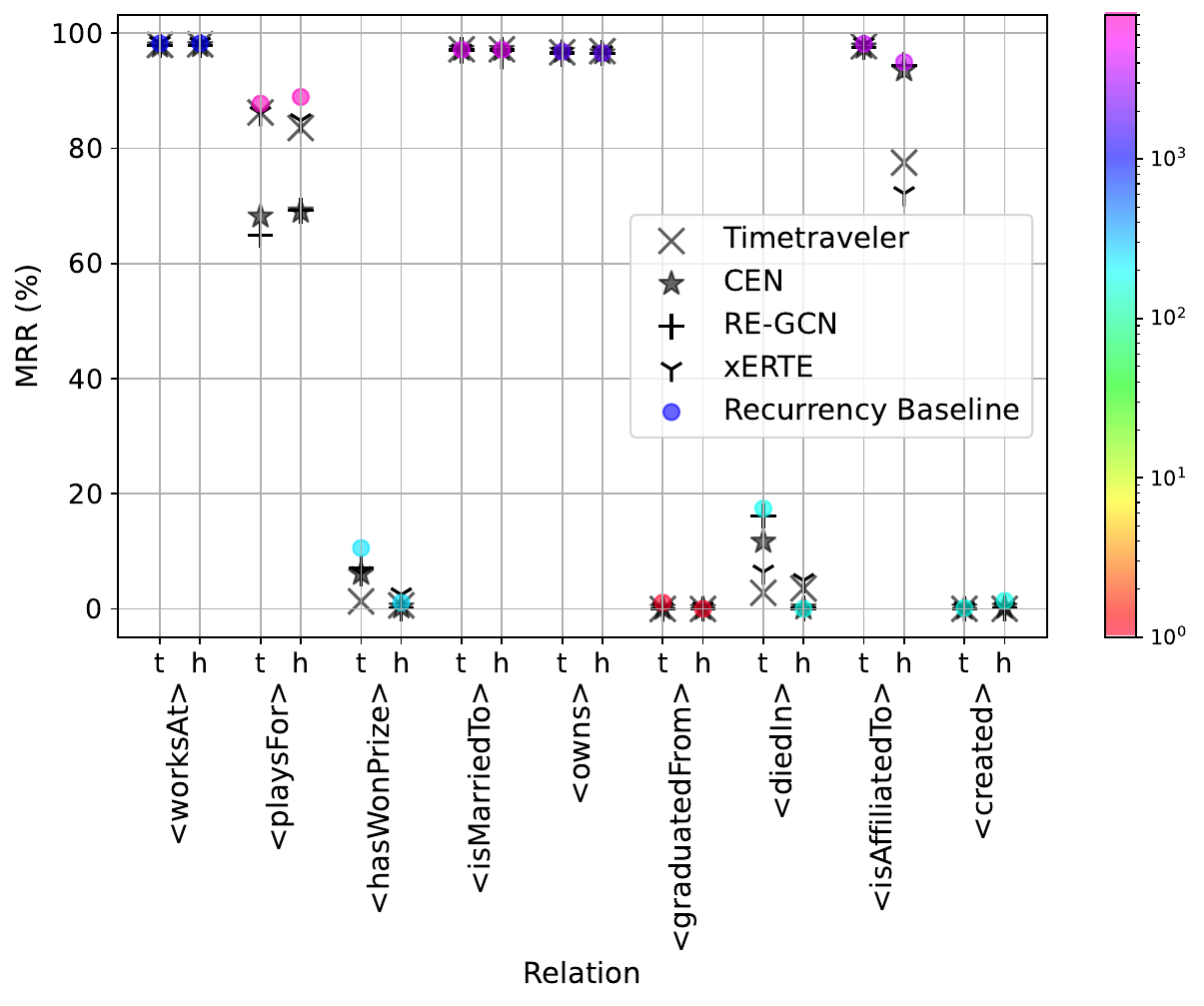}
\end{minipage}
\caption{Test MRRs for each relation and direction (``t'' means tail and ``h'' head, respectively) for (a) ICEWS14 (top) and (b) YAGO (bottom). Colors indicate the number of queries for relation and its direction in the test set.} 
\label{fig:RW}
\end{figure}

\paragraph{ICEWS14}
In Figure~\ref{fig:RW}(a), we focus on the nine most frequent relations.
For each relation, one or multiple methods reach MRRs higher than the \rb, with an absolute offset in MRR of approximately $3\%$ to $7\%$ between the best-performing method and our baseline. This indicates that it might be necessary to capture patterns going beyond the simple recurrency principle. 
However, even for ICEWS14, we see three relations where some methods produce worse results than the \rb. 
For two of these ({Make\_a\_visit}, {Host\_a\_visit}), RE-GCN and CEN attain the lowest MRR. 
In the third relation ({Arrest\_detain\_or\_charge\_with\_legal\_action}), TLogic and xERTE have the lowest MRR. 
This implies that, despite having better aggregated MRRs, the methods display \textit{distinct weaknesses} and are not learning to model recurrency for all relations. 

\paragraph{YAGO} Figure~\ref{fig:RW}(b), instead, shows two distinct categories of relations: 
the first category contains relations where most methods demonstrate competitive performance (MRR$\geq85\%$). In all of them, the \rb attains the highest scores. 
Thus, the capabilities of related work, like detecting patterns across different relations or multiple hops in the KG, do not seem to be beneficial for these relations, and a simpler inductive bias might be preferred.
The second category contains relations where all methods perform poorly (MRR $\leq20\%$).
Due to the dataset's limited information, reliably predicting prize winners or deaths is unfeasible.
For these reasons, we expect no significant improvement in future work on YAGO beyond the results of our baseline. 

However, YAGO still provides value to the research field: it can be used to inspect the methods' capabilities to identify and predict simple recurring facts and, if this is not the case, to pinpoint their deficiencies. Thus, YAGO can be also seen as a dataset for sanity checks. 
All analysed methods from related work fail this sanity check: none of them can exploit the simple recurrency pattern for all relations.
The main disparity in overall MRR between the \rb and related work can be attributed to two specific relations: {playsFor} (head, tail), and {isAffiliatedTo} (head). Queries attributed to these relations make for almost $50\%$ of all test queries.
More specifically, Timetraveler exhibits limitations with {isAffiliatedTo} (head) and {playsFor} (head); xERTE shows its greatest shortcomings for {isAffiliatedTo} (head); and RE-GCN and CEN exhibit limitations with the relation {playsFor} in both directions. These findings highlight the specific weaknesses of each method \textit{that are possible by comparisons with baselines}, thus allowing for targeted improvements.

\subsection{Failure Analysis}
In the following, we analyse some example queries where the recurrency principle offers an unambiguous solution which, however, is not chosen by a specific method.
Following Section~\ref{sec:per-rel}, we focus on YAGO and the same four models. 
We base our analysis on the insights that YAGO has a very high \drec, and that predicting facts based on strict recurrency with steep time decay leads to very high scores. The MRR of $\phi_\Delta$ is $90.7\%$.
For each model, we count for how many queries the following conditions are fulfilled, given the test query $(s,r,?,t)$ with correct answer $o$: (i) $(s,r,o,t\! -\!1) \in G$, (ii) the model proposed $o' \neq o$ as top candidate, (iii) there exists no $k$ with $(s,r,o',k) \in G$. If these are fulfilled, there is strong evidence for $o$ due to recurrency, while $(s,r,o')$ has never been observed in the past. We conduct the same analysis for head queries $(?,r,o,t)$. For each model, we randomly select some of these queries\footnote{
Summing up over head and tail queries for Timetraveler, we find 34 queries that fulfilled all three conditions, for xERTE 149, for CEN 286, and for RE-GCN 525 queries.}
and describe the mistakes made.

\paragraph{Timetraveler} Surprisingly, Timetraveler sometimes suggests top candidates that are incompatible with respect to domain and range of the given relation, even when all above conditions are met. Here are two examples for the "playsFor" (\texttt{pf}) relation, where the proposed candidates are marked with a question mark:
\begin{align*}
& (\ttt{?=spain-national-u23}, \ttt{pf}, \ttt{lierse-sk}, \ttt{10}) \\
& (\ttt{?=baseball-ground}, \ttt{pf}, \ttt{derby-county-fc}, \ttt{10})
\end{align*}
The reasons behind Timetraveler's predictions, despite the availability of reasonable candidates according to the recurrency principle, fall outside the scope of this paper.

\paragraph{xERTE} For xERTE, we detect a very clear pattern that explains the mistakes. In 147 out of 149 cases, xERTE predicts a candidate as subject (object) $c$ when $c$ was given as object (subject). This happens in nearly all cases for the symmetric relation isMarriedTo resulting in the prediction of triples such as $(john,isMarriedTo, john)$. This error pattern bears a striking resemblance to issues observed in the context of non-temporal KG completion in~\cite{meilicke2018fine} where it has already been argued that some models perform surprisingly badly on symmetric relations.

\paragraph{CEN and RE-GCN} Both CEN and RE-GCN exhibit distinct behavior. Errors frequently occur with the "playsFor" relation, particularly in tail prediction. 
In all analysed examples, the types (soccer players and soccer clubs) of the incorrectly predicted candidates were correct. Moreover, we cannot find any other systematic error pattern or explanation for the erroneous predictions. It seems that both models are not able to learn that the playsFor relation follows the simple regularity of strict recurrency, even though this regularity dominates the training set.   

These examples highlight significant insights into the current weaknesses of each method.  Future research can leverage these insights to enhance the affected models. 

\subsection{Parameter Study}\label{sec:res-param}
In the following, we summarize our findings regarding the influence of hyperparameters on baseline predictions. Detailed results are provided in the Supplementary Material.


\subsubsection{Influence of Hyperparameter Values}
We analyze the impact of $\lambda$ and $\alpha$ on overall MRR. Notably, $\lambda$ significantly affects the MRR, e.g., with test results ranging from $12.1\%$ to $23.7\%$ for GDELT across different $\lambda$ values. The optimal $\lambda$ varies across datasets. 
This underlines the influence of time decay: Predicting repetitions of the most recent facts is most beneficial for YAGO and WIKI, while also considering the frequency of previous facts is better for the other datasets. This distinction is also mirrored in the \drec, being notably high for YAGO and WIKI, and thus indicating the importance of the most recent facts. 
Additionally, setting $\alpha$ to a high value ($\alpha \geq 0.99$) yields the best aggregated test results across all datasets, indicating the benefits of emphasizing predictions from the \srb and using the \rrb to resolve ties and rank unseen triples.

\subsubsection{Impact of Relaxed Recurrency Baseline}
Further, to understand the impact of the \rrb ($\xi$) on the combined baseline, we compare the MRR of strict and relaxed baseline on a per-relation basis. 
We find that, even though the aggregated improvement of $\psi_\Delta\xi$ as compared to $\psi_\Delta$ is only marginal ($<1\%$) for each dataset, 
for some relations, where the strict baseline fails, the impact of the relaxed baseline is meaningful: For example, on the dataset YAGO and the relation diedIn (tail), the \srb yields a very low MRR of $0.7\%$, whereas the \rrb yields a MRR of $17.5\%$. 

Overall, this highlights the influence of hyperparameter values, dataset differences, and the advantage of combining baselines on a per-relation basis.

\section{Conclusion}
\label{sec:conclusions}
We are witnessing a notable growth of scientific output in the field of TKG forecasting. 
However, a reliable and rigorous comparison with simple baselines, which can help us distinguish real from fictitious progress, has been missing so far. Inspired by real-world examples, this work filled the current gap by designing an intuitive baseline that exploits the straightforward concept of facts' recurrency. 
In summary, despite its inability to grasp complex dependencies in the data, 
the baseline provides a better or a competitive alternative to existing models on three out of five common benchmarks. 
This result is surprising and raises doubts about the predictive quality of the proposed methods. Once more, it stresses the importance of testing naïve baselines as a key component of any TKG forecasting benchmark:
should a model fail when a baseline succeeds, its predictive capability should be subject to critical scrutiny.
By conducting critical and detailed analyses, we identified limitations of existing models, such as the prediction of incompatible types.
We hope that our work will foster awareness about the necessity of simple baselines in the future evaluation of TKG methods.
\bibliographystyle{named}
\bibliography{references}

\end{document}